\documentclass{vgtc}                          % final (conference style)
\graphicspath{{figures/}{pictures/}{images/}{./}} % where to search for the images

\usepackage{times}                     % we use Times as the main font
\usepackage{tabu}                      % only used for the table example
\usepackage{booktabs}                  % only used for the table example
\usepackage{url}

\usepackage{mathptmx}                  % use matching math font

\usepackage{indentfirst}
\usepackage{subfig}
\usepackage{subcaption}
\usepackage{mwe}% or load ’graphicx’ and ’blindtext’ manually
\usepackage{gensymb}
\usepackage[dvipsnames,table,xcdraw]{xcolor} 
\usepackage[colorinlistoftodos]{todonotes}
\usepackage{tabularx}
\usepackage{color}
\usepackage{soul}
\usepackage{hyperref}
\usepackage{balance}  % PS
\renewcommand{\footnoterule}{   % footnote centered at some point - fix
    \kern -3pt
    \hrule width 0.4\columnwidth height 0.4pt
    \kern 2.6pt
}

\onlineid{0}

\vgtccategory{Research}

\title{How Accurate is the Positioning in VR? Using Motion Capture and Robotics to Compare Positioning Capabilities of Popular VR Headsets}

\author{Adam Banaszczyk\thanks{e-mail: adam.a.banaszczyk@gmail.com} \\
        \scriptsize \small Poznań University of Technology, Poland
\and Mikołaj Łysakowski \\
        \scriptsize \small Poznań University of Technology, Poland
\and Michał R. Nowicki\thanks{e-mail: michal.nowicki@put.poznan.pl} \\
        \scriptsize \small Poznań University of Technology, Poland
\and Piotr Skrzypczyński \\
        \scriptsize \small Poznań University of Technology, Poland
\and Sławomir K. Tadeja\\
        \scriptsize \small University of Cambridge, United Kingdom}

\abstract{
 In this paper, we introduce a new methodology for assessing the positioning accuracy of virtual reality (VR) headsets, utilizing a cooperative industrial robot to simulate user head trajectories in a reproducible manner. We conduct a comprehensive evaluation of two popular VR headsets, i.e., \textit{Meta Quest 2} and \textit{Meta Quest Pro}. Using head movement trajectories captured from realistic VR game scenarios with motion capture, we compared the performance of these headsets in terms of precision and reliability. Our analysis revealed that both devices exhibit high positioning accuracy, with no significant differences between them. These findings may provide insights for developers and researchers seeking to optimize their VR experiences in particular contexts such as manufacturing.
} % end of abstract


\begin{document}


\firstsection{Introduction}

\maketitle

Virtual reality (VR) technology has rapidly evolved, becoming crucial in fields like education and training \cite{Bratyzel_VR_Speech_Training_2024}, engineering design and production \cite{Tadeja_Seshadri_Kristensson_2020}, and healthcare \cite{tadeja_immersive_2024}. Accurate positioning of head-mounted displays (HMDs) is vital for user immersion and interaction in VR \cite{Reimer2021}, ensuring digital objects remain stable and interact correctly with user movements, especially in precision-demanding applications like engineering.

Significant advancements in VR head tracking have focused on improving accuracy, reducing latency, and enhancing user experience \cite{Gourlay2017, LaValle2014}. Developments include low-cost lighthouse-type systems \cite{Ng2017}, outside-in tracking with RGB-D cameras and facial features \cite{Amamra2017}, very accurate visual-inertial tracking \cite{Fang2017}, and large-area headset localization using the Global Navigation Satellite System (GNSS) \cite{Humphreys2020}.

\begin{figure}[tb!]
\centering
\includegraphics[width=0.8\columnwidth]{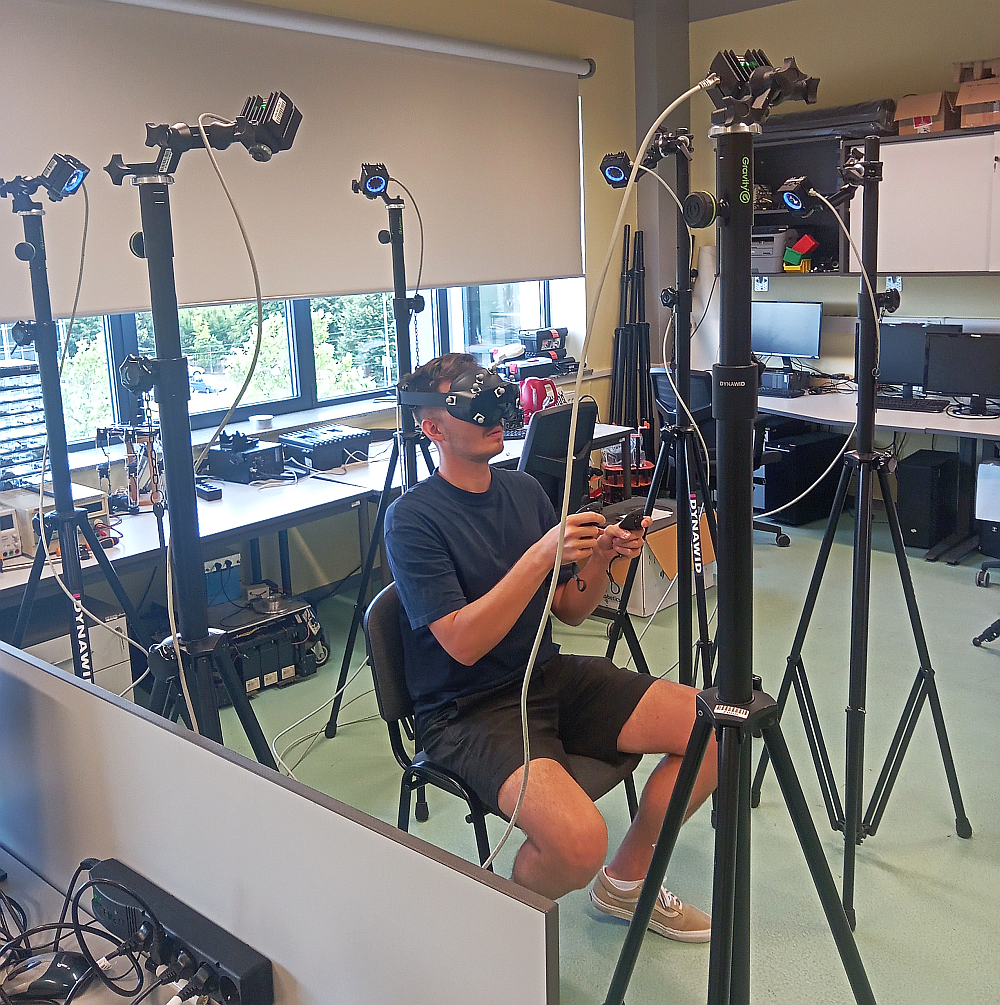}
\caption{Real-world head movement data acquisition with motion capture setup (OptiTrack) during VR gameplay.}
\label{fig:optitrack_gameplay}
\vskip -2mm
\end{figure}

Commercially available VR headsets, like those from Meta \cite{meta2023quest}, use inside-out vision-based positioning, or ego-motion tracking, requiring environmental mapping \cite{Gourlay2017}. These solutions employ visual odometry and simultaneous localization and mapping (SLAM) \cite{Charalambos2022}, demanding expensive computation and struggling in environments with repetitive textures, rapid motion, or significant occlusions \cite{Fu2020}.

Our work presents a new method for evaluating the accuracy of inside-out tracking in novel HMDs, considering possible failure modes. We aim to understand the limits of positioning accuracy in recent VR headsets to guide researchers and developers. Previous research emphasized the importance of precise tracking systems \cite{Ikbal2021, Niehorster2017}, 
as the perception of body position can influence the feel of presence experienced by VR users \cite{Aelee2019}. However, comprehensive evaluations of the latest VR headsets using standardized testing methods for inside-out visual tracking are lacking.

We initiated a project to evaluate the positioning accuracy of VR headsets using real user movement trajectories acquired during VR gameplay. This approach involves various movement sequences, resulting in diverse headset movement trajectories.

For rigorous evaluation, we used an industrial cooperative robot to precisely repeat these head movement trajectories for different headsets. The trajectories, captured using a state-of-the-art motion capture system, serve as the objective ground truth for our measurements (Fig. \ref{fig:optitrack_gameplay}). This ensures consistency and repeatability, allowing accurate headset performance comparison.

We tested two recent mass-market devices: the Meta Quest 2, aimed at the entertainment market, and the Meta Quest Pro, for professional applications. Meta has been the world's largest VR equipment vendor in recent years \cite{XR_vendor_2023}.

Our proposed methodology, software, and data, available at Github\footnote{\texttt{\href{https://github.com/AdamBanaszczyk/head_trajectory}{https://github.com/AdamBanaszczyk/head\_trajectory}}}, will benefit the VR development community. Our contributions include: (i) a detailed comparison of the positioning accuracy of Quest Pro and Quest 2 headsets; (ii) an analysis of VR headsets' tracking precision strengths and weaknesses; and (iii) a reproducible methodology and step-by-step procedure for experiments with various VR headsets using headset-agnostic trajectories, ROS (Robot Operating System) software, and commonly available equipment like a portable motion capture system (OptiTrack) and a cooperative manipulator arm.

\section{Structure of the Positioning Accuracy Assessment System}
The proposed system for assessing VR headset positioning accuracy involves several key components working together for precise evaluations (Fig. \ref{fig:optitrack_gameplay}). The system is structured as follows:

\begin{itemize}
\item Trajectory data collection: this module uses motion capture to collect trajectory data from VR headsets during selected VR applications. It involves calibrating the headset's coordinate system (used by Unity) with the OptiTrack system and synchronizing timestamps between Unity and OptiTrack. The result is real-world trajectories that can be "played" with different VR goggles mounted on a cooperative robot.

\item Trajectory reproduction: the collected trajectories are reproduced by a collaborative robot UR5e with a mock-up head as its end-effector, on which the tested VR headset is mounted. Trajectories are logged simultaneously by the ROS controller and the Unity application. This module requires calibrating the transformation between the robot’s end-effector and the Unity coordinate system. Additionally, trajectories are pre-processed and tested in a Gazebo-based simulation to ensure safe playback by the robot.

\item Assessment of positioning accuracy: this is done using the Absolute Pose Error (APE) from the evo package \cite{grupp2017evo}.  The APE in evo with alignment and translational pose relation is equivalent to the Absolute Trajectory Error (ATE) metric defined in \cite{Endres2012}, and commonly used in robot localization research. The APE aligns both trajectories and measures the distance between the estimated and ground-truth poses, specifically the headset's coordinate system origin as read from the robot and the same point estimated by the VR headset. This requires time synchronization of headset poses from both sources.
\end{itemize}

\begin{figure}[h!]
    \centering
    \includegraphics[width=1.0\columnwidth]{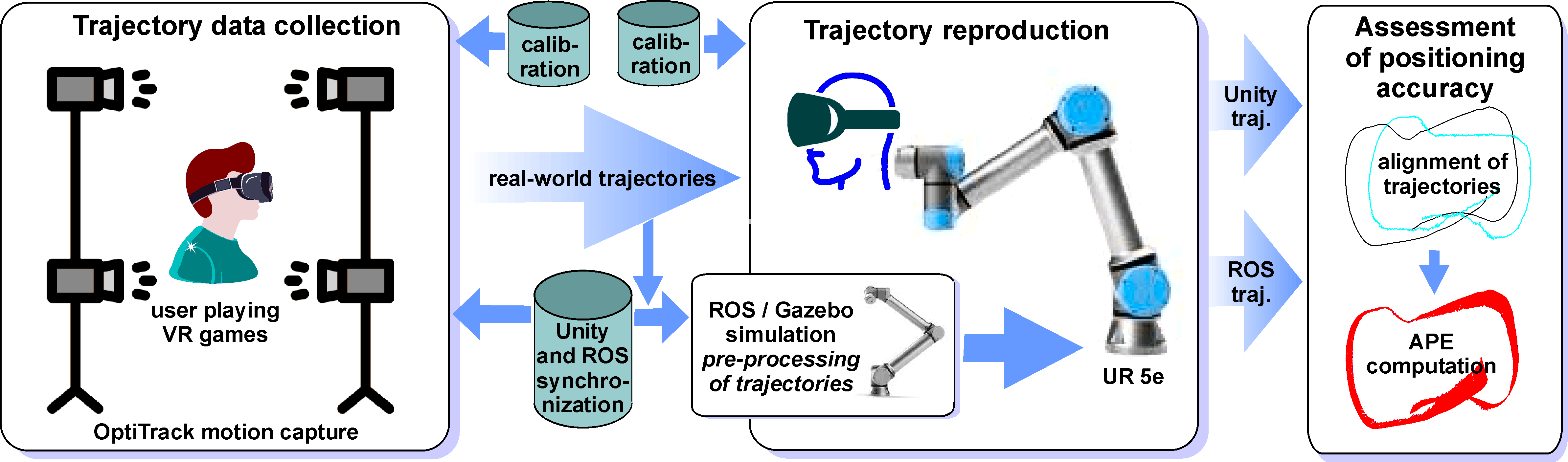}
    \caption{Schematic view of the structure of the proposed positioning accuracy assessment system.}
    \label{fig:system}
\end{figure}

\section{Acquiring Headset Trajectories in Real-World Scenarios}
% \section{Real-World Head Movement Trajectories}
\label{Head_Movement}

\subsection{OptiTrack Motion Capture Setup}
\begin{figure}[h!]
    \centering
    \includegraphics[width=0.99\columnwidth]{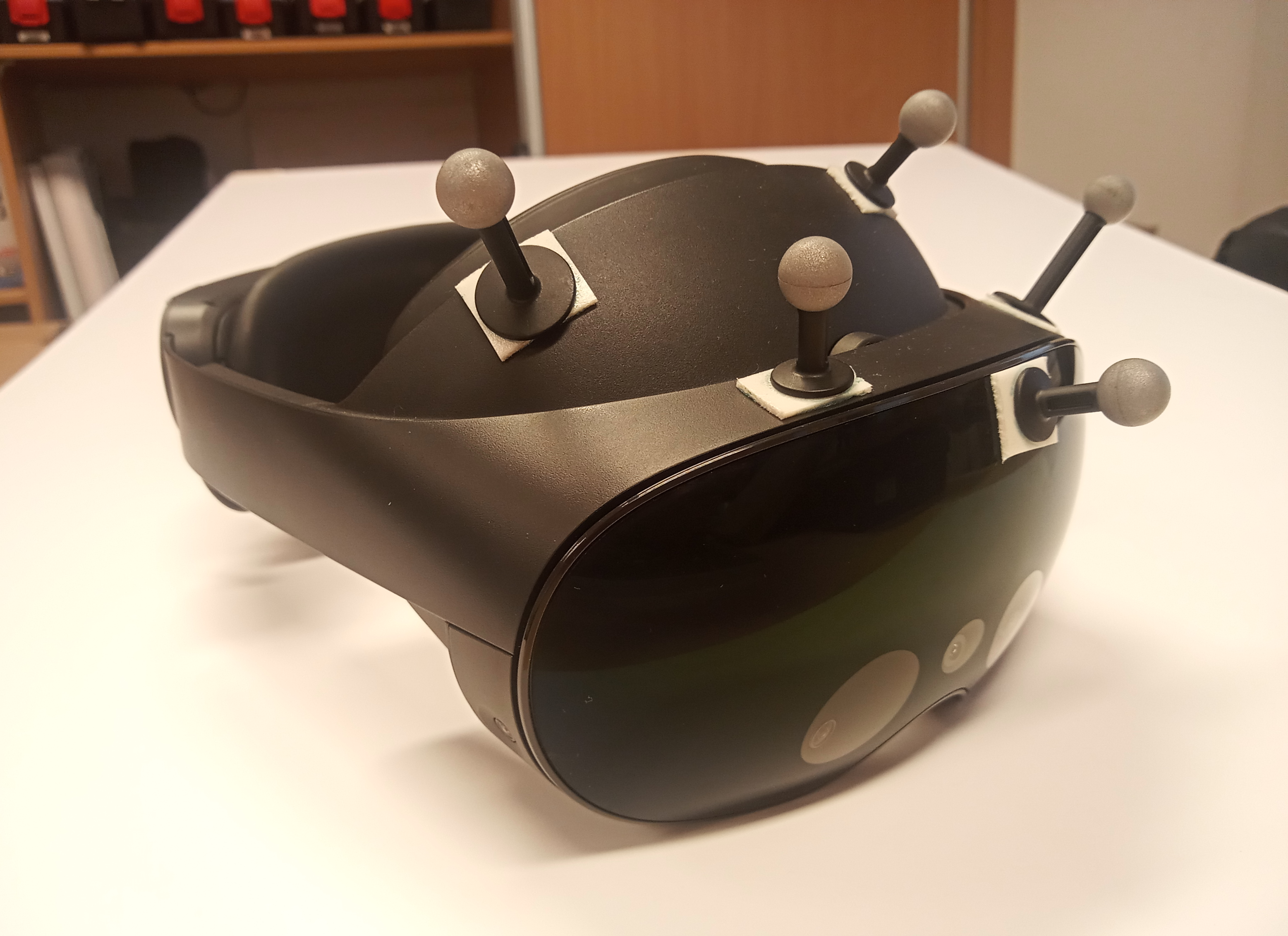}
    \caption{Meta Quest Pro with OptiTrack markers attached.}
    \label{fig:markers}
\end{figure}

To accurately track the user's head motion during HMD usage, we utilized the OptiTrack motion capture system. We started the data capture process by defining a natural playing area for a seated user and surrounding it with six OptiTrack cameras, ensuring that every movement could be accurately recorded (see Fig.~\ref{fig:optitrack_gameplay}).
Each tested headset required a set of OptiTrack markers attached, ensuring no sensor was obstructed (see Fig~\ref{fig:markers}).

% During data acquisition, synchronized cameras captured 2D images and calculated the 2D positions of the markers. Subsequently, overlapping position data was used to triangulate the 3D poses. 
A defined set of markers is later on represented by OptiTrack as a single rigid body corresponding to HMD, with its position defined as the center of mass of used markers. After the calibration procedure, the OptiTrack system tracks the poses of the markers with an expected error smaller than $0.2 [mm]$ at $120[Hz]$ \cite{OptiTrack2019}. 

With the experimental setup complete, the operator could enter the area and proceed to use HMD as usual, allowing for simultaneous trajectory data acquisition. Such a non-invasive setup can be easily replicated with any available VR headset without altering it.

\subsection{Data Acquisition During VR Gameplay}
During the data acquisition, we tested a pool of VR games and experiences based on their compatibility with a defined OptiTrack's playing area, accessibility, popularity, and user ratings. Each tested app had to be free or available in a playable demo version, suitable for seated gameplay, rated more than 1,000 times, and have a user rating of at least 80\% positive. We tested, among others, \textit{Space Pirate Trainer DX}, \textit{Oculus First Contact}, \textit{Bait!}, \textit{Mission: ISS: Quest}, and \textit{Gorilla Tag}.

\begin{table}[h]
\begin{tabularx}{\columnwidth}{Xcc}
\hline
\textbf{Game} & \textbf{Score} & \textbf{Ratings} \\
\hline
\textit{Space Pirate Trainer DX}  & 92\% & 3498 \\
\textit{Oculus First Contact}     & 92\% & 1451 \\
\textit{Bait!}                    & 81\% & 2987  \\
\textit{Mission: ISS: Quest}      & 80\% & 1619 \\
\textit{Gorilla Tag}              & 93\% & 120k+ \\

\hline
\end{tabularx}
\caption{Statistics of tested applications as obtained from \href{https://vrdb.app/quest/index_us.html}{\textit{https://vrdb.app/quest/index\_us.html}}, where Score indicates the average user rating for the application, while Ratings represent the total number of votes cast by users who have purchased or downloaded the app.}
\label{table:games}
\end{table}

While using a headset, we collected 6 degrees of freedom (DoF) positional ground-truth data in the form of Robot Operating System (ROS) messages with the help of OptiTrack. We chose ROS format for its standardized structure and Unix timestamps, guaranteeing uniformity in data comparison across all experiments. However, the OptiTrack software (Motive) does not directly support the ROS format and only measures time increments from the initial entry, which was insufficient for our application. Hence, we created a custom package based on the NatNet client/server networking software development kit (SDK), which allowed us to forward Motive messages to another computer publishing ROS messages. To eliminate the transfer lag that potentially could have disturbed the direct comparison of OptiTrack and HMD data, we connected the computers directly via Ethernet.

\subsection{Setup Calibration: Unity-to-OptiTrack}
Having precisely recorded real-world head trajectories in the OptiTrack coordinate system, we needed to determine its exact transformation to the respective trajectory recorded by the HMD in its coordinate system. To obtain such data, we first created a Unity application, which at the frequency of $30[Hz]$ logged the device's timestamps together with 6 DoF quaternion poses of the \texttt{Center Eye Anchor}, provided by the Oculus Software Development Kit (SDK) and the Oculus Integration Package, which manages the VR camera setup and tracks its alignment with the physical movements. \texttt{Center Eye Anchor} references a single point in the VR environment representing the position and orientation of the user's eyes or head. Hence, it was well-suited for our needs.

In the course of all the described experiments with the Meta Quest Pro and Meta Quest 2 headsets, the software version was Meta Quest build $60.0.0$, and the OS version was $SQ3A.220605.009.A1$.

The calibration procedure consisted of a set of specifically prepared relatively slow HMD movements, which aimed at proper saturation of poses, enabling accurate calculation of transformation between two systems. The procedure was performed in the arranged playing area, simultaneously recording data from both introduced systems (OptiTrack and Unity). It was crucial to maintain the identical configuration of OptiTrack markers as utilized during the actual real-world head trajectory recordings since any discrepancies would pollute the transformation.

Determining the precise transformation between Unity's Center Eye Anchor and OptiTrack's rigid body required solving a hand-eye calibration problem \cite{Hand-Eye_Horaud_1995}. Both systems were rigidly attached to the same actuator, which in this case was the user's head. Therefore, there existed a homogenous transformation $X$ that we aimed to determine that satisfies the equation $AX = XB$, where $A$ denotes the transformation from the Unity base to the Center Eye Anchor, and $B$ represents the transformation between the previous and current poses of the OptiTrack rigid body whenever the system moves. To resolve this problem, we developed a C++ program utilizing the OpenCV~\cite{opencv_library} and Eigen~\cite{eigenweb} libraries, which sequentially loaded both trajectories, applied linear interpolation for translation and spherical linear interpolation (slerp) \cite{slerp} for quaternion rotation, matched their entries by timestamps, and transformed Unity's trajectory from a left-handed system (LHS) into the right-handed system (RHS). The latter was chosen as it matches OptiTrack, ROS, and the majority of robotics applications. 
% To change from LHS to RHS, we negated the direction of the Y axis, which required negating ${y, qy, qw}$ components of the 6 DoF quaternion representation. 
Each sequence had its data represented as a transformation from the starting position. Moreover, since OptiTrack's frequency was 4 times greater than Unity's, we diluted its data by that factor to significantly increase the speed of code execution. 
% We thoroughly tested the existing procedures for that calibration in OpenCV and found that Park \cite{Park_Hand-Eye}, Horaud \cite{Hand-Eye_Horaud_1995}, Andreff \cite{Andreff_2001} and Daniilidis \cite{Daniilidis_Hand-Eye} methods solved the desired transformation precisely in all studied cases, while Tsai \cite{Tsai_Hand-Eye} method struggled. Thus, in all the experiments that followed, the Daniilidis method was used.
% We thoroughly evaluated the available calibration methods in the OpenCV library and selected the Daniilidis \cite{Daniilidis_Hand-Eye} method for our experiments due to its consistent performance. Additionally, we found that the Tsai \cite{Tsai_Hand-Eye} method occasionally failed to yield accurate solutions.
We thoroughly evaluated the available calibration methods in the OpenCV library, and we have found that 
the Tsai \cite{Tsai_Hand-Eye} method sometimes failed to provide accurate solutions, so we selected the Daniilidis method \cite{Daniilidis_Hand-Eye} due to its consistent performance.

%Unity-to-OptiTrack transformation allows for the depiction of each system in a common reference frame, thereby making it possible to directly compare trajectories and calculate any deviations or errors.

During hand-eye calibration 
% we encountered an unexpected issue that accompanied most of the experiments concerning Unity's trajectories. While comparing its timestamps with the ones recorded by the OptiTrack
we discovered a slight time offset between OptiTrack's and Unity's trajectories, which varied for different recordings but was within $0-200 [ms]$. We tried to eliminate that timestamp mismatch by ensuring that HMD is fully charged and OptiTrack's computer and HMD are connected to the Internet to synchronize internal clocks. However, a slight timestamp offset was still present in some cases. Therefore, we needed to introduce an additional function that could estimate the value of that delay with an accuracy of a few milliseconds. This new program detected axial position peaks for both trajectories and compared the respective timestamps throughout entire recordings. Differences in timestamps at those particular positions were then averaged to calculate the final time offset between Unity and OptiTrack, which was addressed during the hand-eye calibration and any other experiment, allowing for precise time alignment of studied trajectories.

\section{Mounting Headset on The Robotic Arm}
\subsection{Custom Robot Mount}
\label{Installation}

% \begin{figure}[h]
% \includegraphics[angle=-90, width=0.236\textwidth]{figures/mount1.jpg}\hfill
% \includegraphics[angle=-90, width=0.236\textwidth]{figures/mount2.jpg}
% \caption{Artificial head with the Meta Quest Pro attached, mounted on the Universal Robots UR5e manipulator tip}
% \label{robot_mount}
% \end{figure}
\begin{figure}[h!]
    \centering
    \includegraphics[width=0.94\columnwidth]{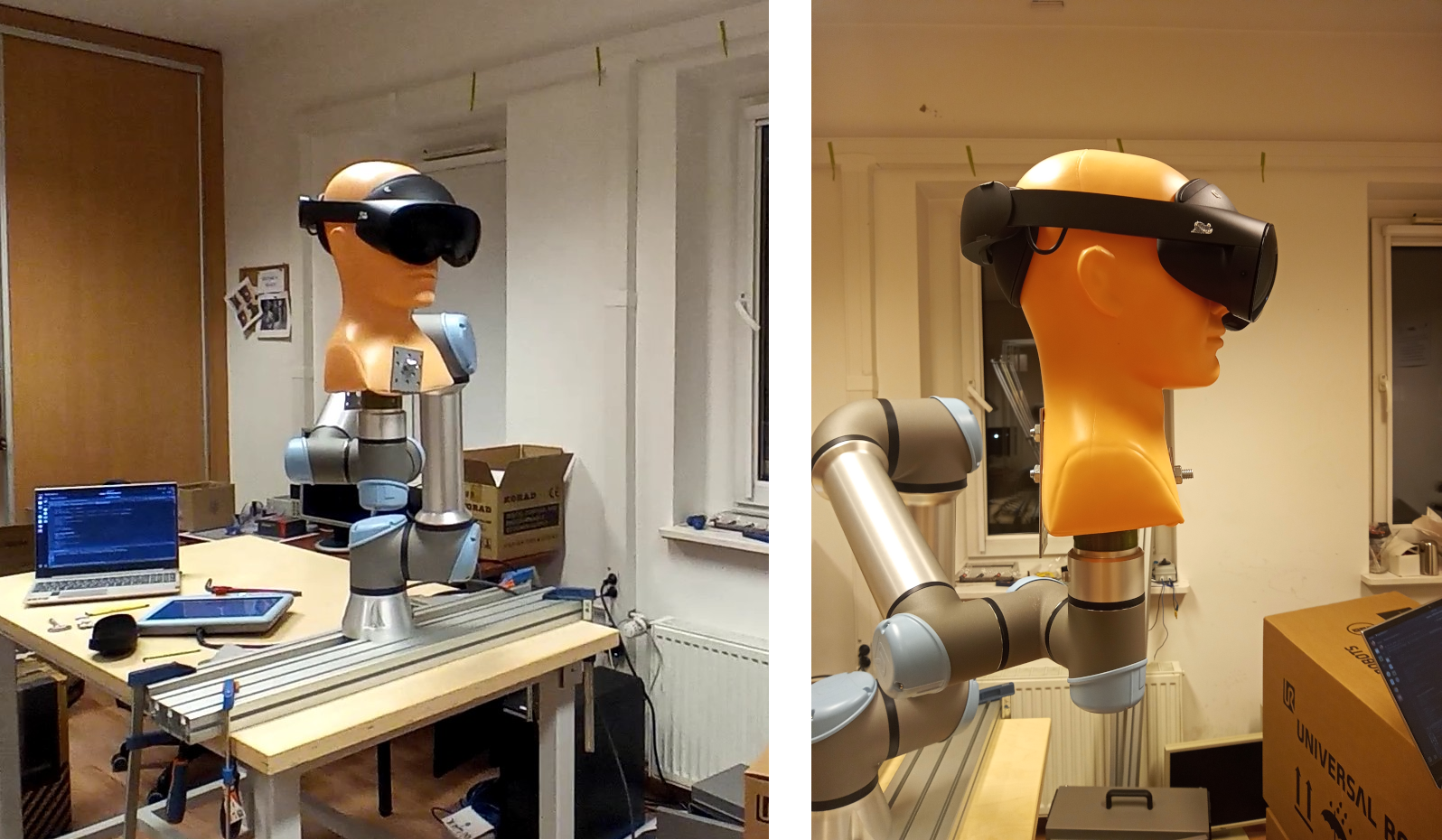}
    \includegraphics[width=0.94\columnwidth]{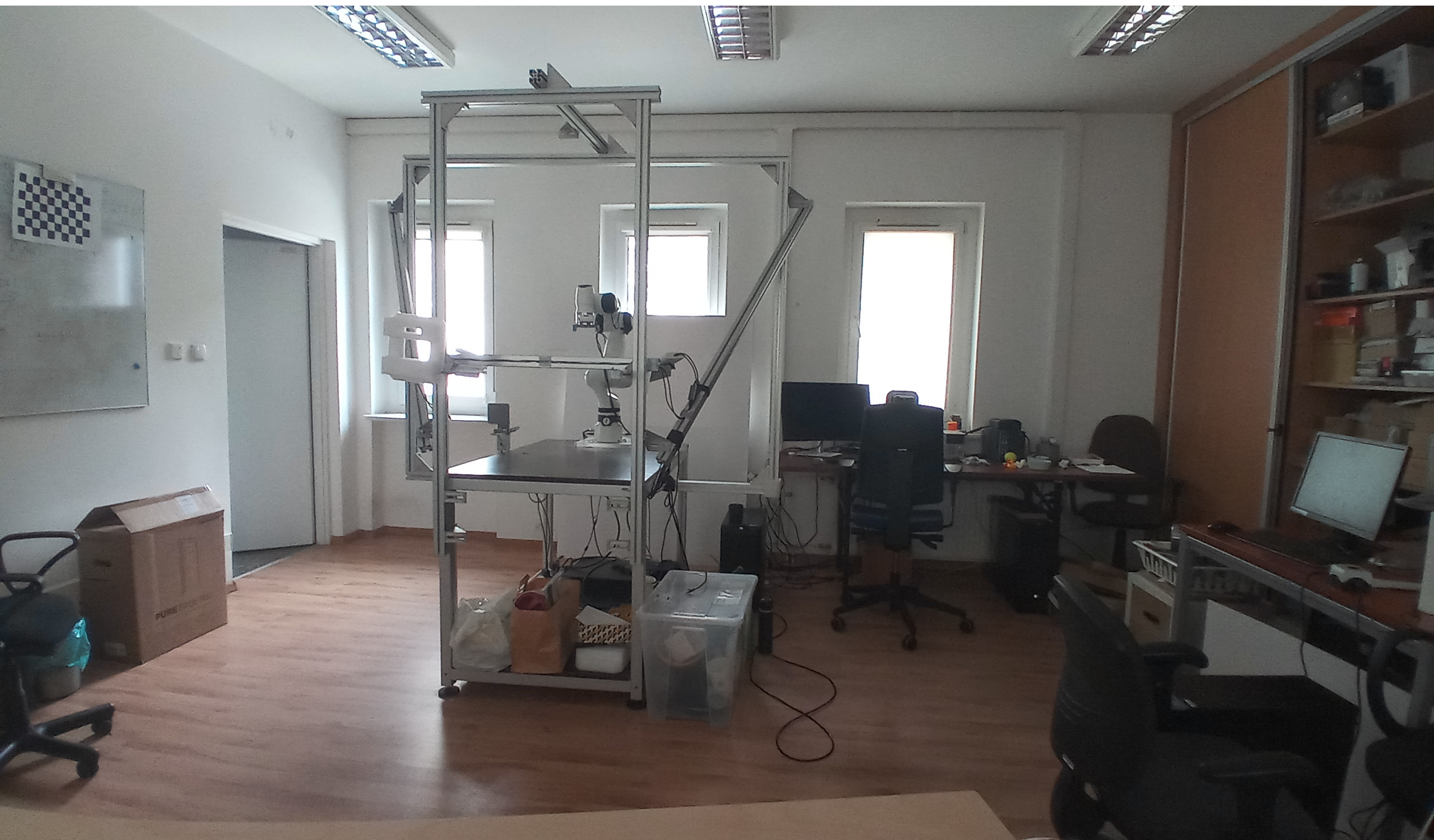}
    \caption{Artificial head with the Meta Quest Pro attached, mounted on the Universal Robots UR5e manipulator tip (up), and the real-world test environment as seen from the perspective of the mounted headset (down).}
    \label{fig:robot_mount}
    \vskip -2mm
\end{figure}

Accurate replication of human head movements on a robot requires a machine capable of performing demanding movements with high accuracy and precision of repetition, while having a relatively large range and freedom of motion to guarantee the ability to reproduce all sorts of trajectories. Consequently, we decided to use Universal Robots UR5e robotic arm with a maximum reach of $850 [mm]$, payload of up to $5[kg]$, six rotating joints, and pose repeatability of $\pm0.03[mm]$~\cite{ur}.
Accuracy is understood to be the ability to attain the required point in the workspace. While this parameter is not directly specified by the robot manufacturer (as repeatability is mainly important in industrial tasks), results reported in the literature \cite{Pollak2020} show that the unidirectional pose accuracy in UR5 is better than 0.1 [mm], under a combination of two load values and three motion speeds (up to 100\% speed). This suggests that the UR5 robot is suitable to replicate the headset's motion in our experiments. 

The robot itself was mounted on a dedicated table, resulting in the average height of the Tool Center Point (TCP) within the height of a human head when playing in VR. Although the UR5e robotic arm is very versatile, we needed to propose a customized, secure and robust HMD mounting. Therefore, as all VR headsets are designed to ergonomically rest on the user's head, which is a uniquely shaped body part itself, we decided that the optimal approach was to place the headset on a plastic mannequin bust (see Fig.~\ref{fig:robot_mount}). Furthermore, we designed a custom mounting system to connect such a head to the UR5e manipulator tip, which features threaded holes into which the mounting system was securely fastened. The mounting system consisted of 3D-printed components responsible for precise alignment and connection of all elements and sturdy steel plates providing stability to the base of the mannequin bust (see Fig.~\ref{fig:robot_mount}).

\subsection{Setup Calibration: TCP-to-Unity}
\label{TCP-to-Unity}
With HMD properly mounted on the robot, we were able to determine another crucial transformation between the robot's TCP expressed in the robot's global frame, with the coordinate system origin at its base and Unity's Center Eye Anchor recorded with our application running on the headset. Designating accurate TCP poses in the established earlier format, i.e., 6DoF positions with quaternion rotations and timestamps, required setting up communication with the robot's controller via Real-Time Data Exchange (RTDE) client library, which implements API for UR interface. We eliminated potential transfer delays by connecting an external computer directly to the robot's controller via Ethernet, which ensured that read timestamps were accurate. 

For convenience, we created a ROS publisher for the TCP poses, ensuring the data was in the same format as the OptiTrack output so all the scripts could read it. 
%UR5e also uses a RHS. However,
In addition, the RTDE read TCP's rotation, which had to be converted from the Rodrigues' rotation vector (Angle-Axis representation) to the quaternion representation used in typical ROS messages. We set the frequency of the publisher to $30 [Hz]$, matching the value used by the Unity system.

Calibration movement for this task was again designed to properly saturate different headset poses. This time, the process was automated with a script that contained a set of TCP poses the robot traversed with slow and steady joint movements. The commands were uploaded to the controller with the RTDE library, and during the movement, we recorded both Unity and TCP positional data. It is worth mentioning that in order to have the opportunity to fully control the headsets, whenever they were attached to the mannequin head, we used the casting option, which allowed us to stream the HMD display onto an external device, such as a computer or a smartphone. This was crucial for controlling running apps without taking the headset off and disrupting or altering the values of any transformations. 

With proper data recorded, we once again solved the hand-eye calibration problem \cite{Hand-Eye_Horaud_1995} identically to our previous approach. Since the Unity application ran on the HMD, we again needed to address potential time offsets between Unity and the UR5e. This offset was calculated using the peak detection method described earlier.

\section{Determining Robot Motions}
\subsection{Robotic Arm Movement Simulator}
Having calculated all the required transformations, we could now determine the precise movements of the robot needed to reproduce any HMD's real-world trajectory that we previously recorded. This necessitated the creation of a simulation environment where various robotic experiments could be safely conducted. We created such a dedicated Docker ROS environment with the following set of essential packages:
\texttt{UniversalRobots/Universal\_Robots\_ROS\_Driver}, that was responsible for creating a general interface between UR robots and ROS;
\texttt{ros-industrial/universal\_robot}, which provided ROS nodes for communication with UR5e controller;
\texttt{MoveIt} \cite{MoveIt}, a robotics motion planning framework.

% \begin{itemize}
%    \item \texttt{UniversalRobots/Universal\_Robots\_ROS\_Driver}, that was responsible for creating a general interface between UR robots and ROS;
%    \item \texttt{ros-industrial/universal\_robot}, which provided ROS nodes for communication with UR5e controller;
%    \item \texttt{MoveIt} \cite{MoveIt}, a robotics motion planning framework.
% \end{itemize}

To test the robot's movements, we first had to bring up the exact model of UR5e in the Gazebo \cite{Gazebo} simulator with a pre-made launch file. Next, we integrated it with MoveIt, which enabled the control of the model and the initiation of motion planning. We also configured UR5e in RViz, a 3D visualizer for the ROS framework, allowing for free manipulation of the simulated robot, which was convenient for setting up different starting poses.

\subsection{Calculating TCP Poses and Robot's Movements}
Following developing an accurate and responsive simulator, we calculated the successive TCP poses based on the real-world head trajectory and verified their sequential executability, ensuring the trajectory's accurate reproduction. 

Firstly, we loaded the head trajectory and diluted the number of individual poses %by a factor of ten
since OptiTrack recorded the data at $120[Hz]$, which was an excessive amount to reproduce for a robot causing undesired vibration during UR5e's movement. The value of this dilution parameter was experimentally fitted and resulted in 12 poses per second. Thus, striking a good balance between accurately reproducing the trajectory and minimizing robot oscillations.

Next, we defined the initial TCP pose, as it directly influenced the entire motion sequence, where each pose had to incrementally follow the calculated movement to accurately replicate the movement of the HMD's Center Eye Anchor. The correct starting pose, expressed in the robot's RHS, also needed to be determined for each studied trajectory to help prevent self-collision and to ensure the robot did not attempt to reach positions outside its operational range. Consequently, each OptiTrack pose was converted into an increment relative to the first pose of its respective trajectory. 

\begin{figure}[h!]
    \centering
    \includegraphics[width=0.95\columnwidth]{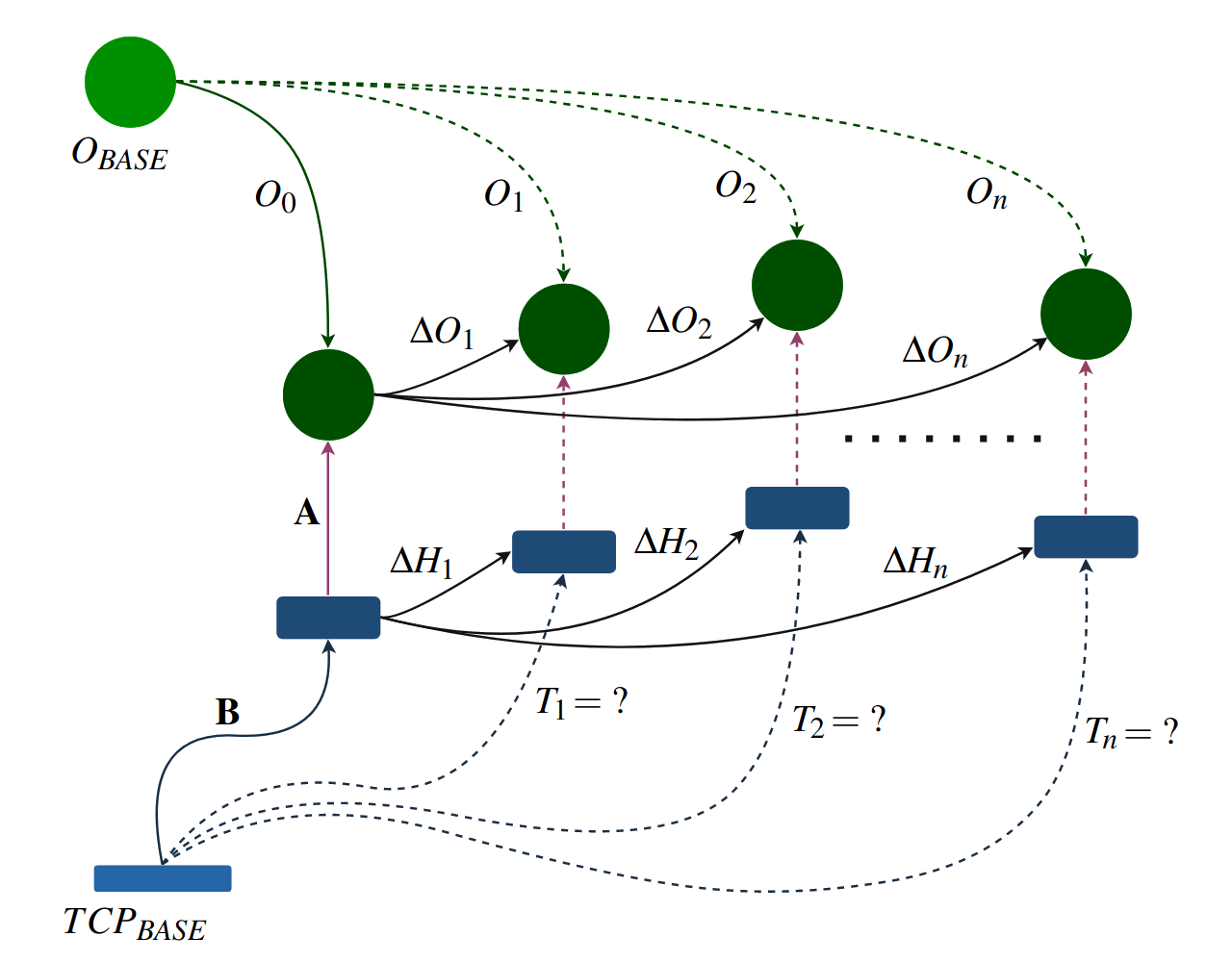}
    \caption{Diagram illustrating the relation between systems when determining TCP poses of interest $T_i$ for each OptiTrack pose of the real-world head trajectory $O_i$. The diagram consists of two fixed elements: $\mathbf{A}={^{\rm TCP}}\mathbf{T}_{\rm Unity} \cdot {^{\rm Unity}}\mathbf{T}_{\rm Opti}$, representing the transformation between the TCP and OptiTrack's Rigid Body corresponding to the VR headset, and $\mathbf{B}$, which denotes the initial pose of the TCP in its coordinate system. $\Delta{O}_{i}$ and $\Delta{H}_{i}$ represent the changes in poses within their respective systems, relative to the starting point, for the trajectory recorded by OptiTrack and the target TCP trajectory. The final TCP poses of interest can be determined with the formula $T_i = \mathbf{B} \cdot \Delta{H}_{i}$, where $\Delta{H}_i=\mathbf{A} \cdot \Delta{O}_i \cdot \mathbf{A}^{-1}$.}
    \label{fig:transforms_explained}
\end{figure}

Our approach ensured consistency in HMD movement in its independent system, regardless of the chosen starting pose. With this in mind, in order to calculate a set of UR5e's TCP poses of interest for each OptiTrack pose of real-world head trajectory, we needed to solve the following problem, depicted in Fig.~\ref{fig:transforms_explained}:
\begin{equation}
    T_i = \mathbf{B} \cdot {^{\rm TCP}}\mathbf{T}_{\rm Unity} \cdot {^{\rm Unity}}\mathbf{T}_{\rm Opti} \cdot {\Delta{O}_i} \cdot {^{\rm Opti}}\mathbf{T}_{\rm Unity} \cdot {^{\rm Unity}}\mathbf{T}_{\rm TCP}
\end{equation}
where ~$\mathbf{B}$ is an aforementioned initial pose of TCP in the robot's coordinate system with an origin at its base, each~$\mathbf{T}$ refers to a particular transformation matrix and ~${\Delta{O}_i}$ is an increment of a pose at given index in real-world head trajectory recorded with OptiTrack in its respective coordinate system relative to the starting pose, which can be described as:
\begin{equation}
    {\Delta{O}_i} = {{O}_{0}}^{-1} \cdot {O}_i
\end{equation}
The initial increment in the OptiTrack's head pose is a special case, where the given delta is simply the identity matrix ~${\Delta{O}_0 = \mathbf{I} }$

With the target TCP poses calculated, we determined the corresponding joint positions necessary for accurately completing the given trajectory within the MoveIt framework. This process defined the final robot movement path.
%At this stage, we adjusted the parameter responsible for the maximum error in reproducing each TCP pose as the initial base value of $eef\_step = 1[cm]$ resulted in visible oscillations of the robot's movements. This $eef\_step$ parameter refers to a maximum allowable misalignment of TCP pose.
At this stage, we adjusted the value of the $eef\_step$ parameter, which refers to the maximum allowable misalignment of TCP pose during trajectory reproduction in a simulated environment, because its base value of  $eef\_step = 1[cm]$ caused the robot to oscillate during some challenging movements.

The most critical factor at this point was to ensure the exact replication of the robot's movements each time a given sequence was triggered while eliminating any vibrations or similar disturbances. To achieve this, we reduced the precision required for passing through each TCP pose to $eef\_step = 3[cm]$, which still resulted in the robot performing a movement analogous to the VR user's head, at the same time guaranteeing the fluidity of every motion. Consequently, in the vast majority of cases, the reproduction of TCP pose was still extremely accurate. Only a few specific cases caused by unfavorable robot joint alignments were problematic and necessitated increasing the value of the $eef\_step$ parameter during MoveIt motion planning and simulation.

%It is also worth mentioning that \textit{eef\_step} refers to a maximum allowable misalignment of TCP pose, and in the vast majority of cases, the reproduction of TCP pose was still extremely accurate. Only a few specific cases caused by unfavorable robot joint alignments were problematic and influenced the alteration of the parameter.

As a result of the simulation, we obtained the robot's final path together with the corresponding $fraction$ value, which describes the extent to which the robot successfully reproduced the TCP poses for the given parameters and the indicated starting pose. Whenever the robot self-collided, tried to reach a point outside its working area, or could not find a direct movement between following trajectory points, the $fraction$ value dropped significantly. Thus, the critical task was to find such a configuration in which the ~$fraction = 1.0$, which meant that all of the defined poses were achieved. %Consequently, only for such cases did we accept the robot's path and proceed with the experiments.
% It required approximately a dozen attempts
Finding such a configuration per single trajectory required several dozen attempts in the simulation, and only in these cases did we accept the robot's path and proceed with the experiments.

\section{Robot Reproduced Real-World Trajectories}
The testing proceeding was identical for each VR headset, beginning with mounting the device on the robot in the same manner as described in Section \ref{Installation}. This was followed by the calibration procedure detailed in Section \ref{TCP-to-Unity}, which provided accurate TCP-to-Unity transformation. 

After completing the above activities, we immediately proceeded to replay the robot's trajectory obtained from the simulation, resulting in a real-world head trajectory in a test environment on the robot. We accomplished this with the assistance of the RTDE library, which directly sent commands containing a set of target joint positions to the robot's controller. 

During such a test, we recorded Unity's Center Eye Anchor trajectory with the help of our application installed on the headset and UR5e TCP trajectory with our ROS publisher. It was crucial not to disturb the headset's position between the calibration process and the final test to ensure accurate calculation of the results. Hence, we remotely controlled the headset by enabling the casting option during all the experiments to eliminate any need to touch or adjust the headset.

\section{Headsets Positioning and Accuracy}
For the final evaluation and comparison of trajectories recorded with different headsets, we used evo \cite{grupp2017evo}, which is a Python package specifically designed for such tasks.

To use this package, every studied trajectory needed to be depicted in a compatible format,
%therefore we saved them as TUM-formatted text files. In this format, each row contains 8 entries: timestamps, position, and quaternion orientation, all separated by a space.
therefore we saved them as text files, where each row contains 8 entries: $timestamp\ x\ y\ z\ q_{x}\ q_{y}\ q_{z}\ q_{w}$ corresponding to timestamps, position, and quaternion orientation, all separated by a space. This trajectory format, first introduced in \cite{Endres2012}, is known as TUM.

Before saving trajectories in the final TUM-formatted text files, they all needed to be converted to a common coordinate system, and their timestamps had to correspond. We started with converting Unity's trajectory to the RHS by flipping the $Y-axis$, which required negating ${y, q_{y}, q_{w}}$ components. 

Subsequently, as the ground truth during the final tests was the TCP trajectory in the robot's coordinate system, accurately recorded by its controller, we needed to transform each pose from the HMD's Unity trajectory into that system. Essential to this process was the TCP-to-Unity transformation established at the beginning of the final test. We also needed to ensure that all evaluated trajectories had an identical starting pose so that we could align them with the origin of the robot's coordinate system. The definitive form of compared trajectories ${GT, P}$, saved to TUM format, could be described with the following equations:
\begin{equation}
    {GT}_i = \mathbf{X}^{-1} \cdot {\rm TCP}_i
\end{equation}
where ~${GT}_i$ is a final ground truth pose at index \textit{i}, ~$\mathbf{X}$ is a transformation from the origin of the robot's system to the first pose of a TCP trajectory, and ~${\rm TCP}_i$ is a pose from TCP trajectory at index \textit{i}.

\begin{equation}
    {P}_i =  {{^{\rm Unity}}\mathbf{T}_{\rm TCP} \cdot \mathbf{Y}^{-1} \cdot {Unity}_i \cdot ^{\rm TCP}}\mathbf{T}_{\rm Unity}
\end{equation}
where ~${P}_i$ is a pose from the final HMD Unity's trajectory transformed to the robot's system aligned with its origin, ~$\mathbf{Y}$ is a transformation from the origin of the Unity's system to the first position of its trajectory, and ~${^{\rm Unity}}\mathbf{T}_{\rm TCP}$ is a TCP-to-Unity transformation, while ~${^{\rm TCP}}\mathbf{T}_{\rm Unity}$ represents its inverse.
For the timestamps, we employed the peak-detection method to synchronize all trajectories to matching periods.

\subsection{Evaluation of Calibration Accuracy}
We first needed to check how precise were the trajectory reproductions in the non-experimental setup, such as the robot's calibration movement described in Section \ref{TCP-to-Unity}. Therefore, we compared two separate yet identical executions of the specified robot with different headsets mounted on the robot, namely the Meta Quest 2 and Quest Pro. During each execution, we collected TCP and HMD Unity trajectories. As for the results of their respective TCP-to-Unity hand-eye calibration, we obtained a satisfactory transformation with $9.3~[mm]$ translation and $0.178\degree$ rotation error for Quest 2 and $6.2[mm]$ translation $0.209\degree$ rotation error for Quest Pro.

With the necessary transformations calculated, we could now precisely evaluate the performance of the headsets and UR5e during calibration. In the presented results, we removed sections of the idle trajectory before initiating the robot's movement and cut the trajectory just before the final brake, which caused the mounted headset to vibrate significantly.

\begin{figure}[h!]
    \centering
    \includegraphics[width=0.98\columnwidth]{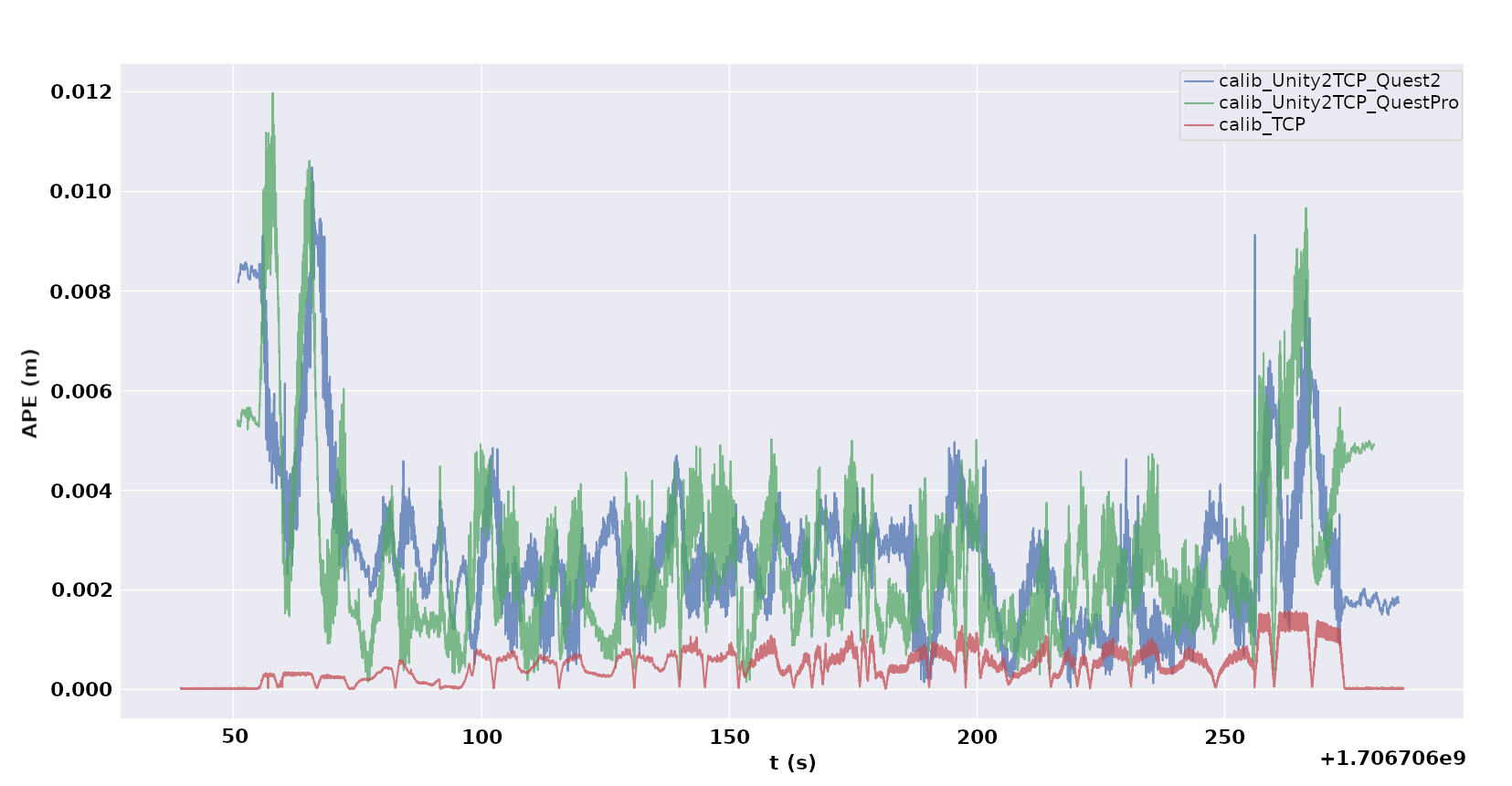}
    \includegraphics[width=0.98\columnwidth]{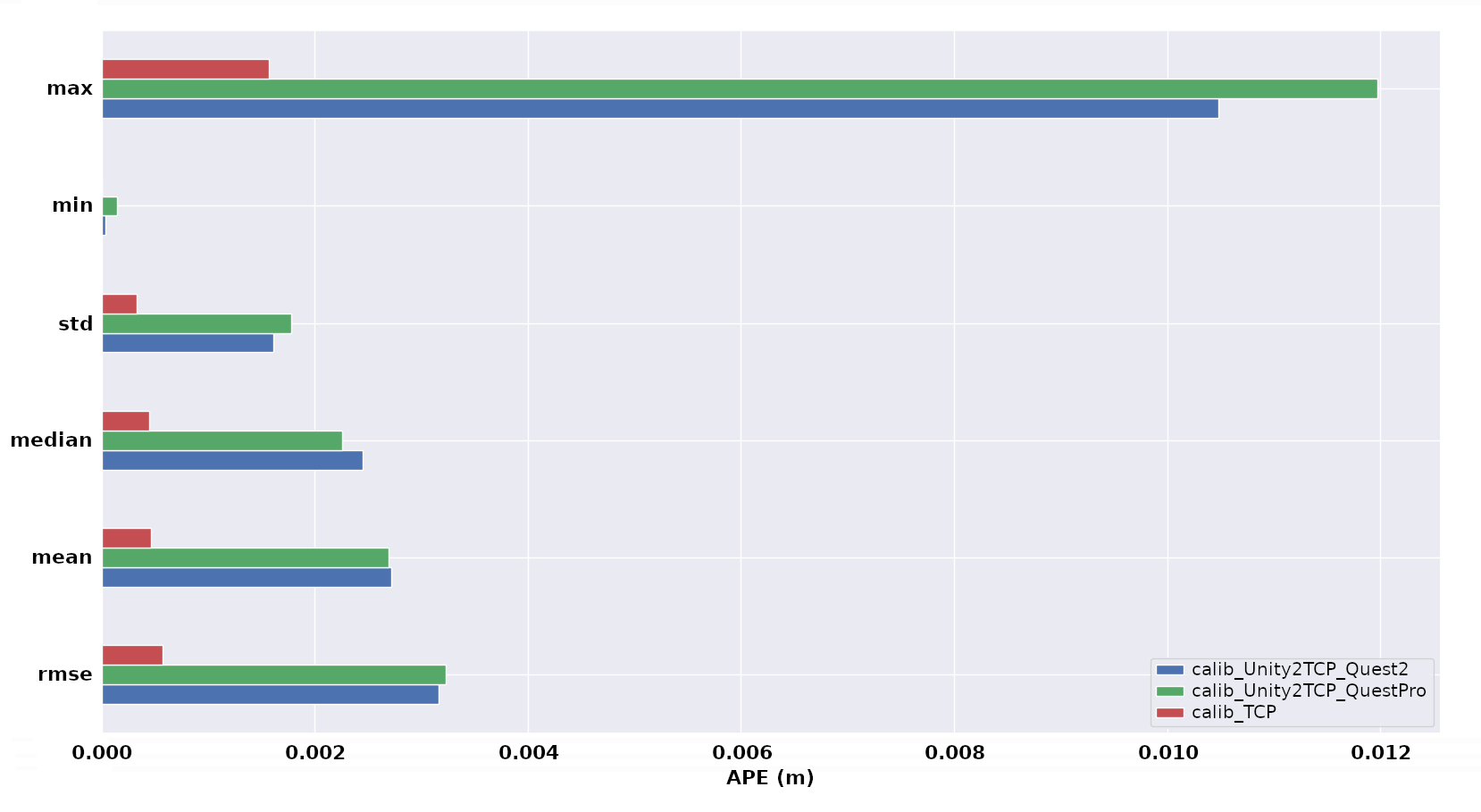}
    \caption{APE plot (up) and diagram (down) for Quest 2, Quest Pro, and TCP trajectories between two identical executions of the robot's calibration motion, also comparing TCP poses during these movements in red. The abscissa displays Unix timestamps (i.e. no. of seconds since Jan 1, 1970), represented in scientific notation.}% corresponding to the number of seconds since January 1, 1970.}
    \label{fig:APE_calib}
\end{figure}

The APE for TCP trajectories, meaning the pose differences recorded by the robot's controller during those calibration motions averaged at $0.467[mm]$ with root mean square error (RMSE) of $0.057[mm]$. These results prove the extreme repeatability of the UR5e robotic arm together with a reliable maximum error of only $1.5[mm]$. The magnitude of this error aligns closely with the accuracy specified by the manufacturer, therefore we regarded these results as acceptable.

In the case of the HMDs accuracy, we obtained $3.17[mm]$ and $3.23[mm]$ of RMSE for Quest 2 and Quest Pro, respectively, with max errors of $10.48[mm]$ and $11.97[mm]$, as shown in Fig.~\ref{fig:APE_calib}. Given that achieving these results for each headset required the use of pre-calculated transformations, the outcomes can be regarded as accurate within the specified environment, with pose tracking being nearly identical for both VR devices.

It is important to note that the volume of the calibration trajectory was approximately $0.3~m^3$, whereas the volume of the trajectory for real-world headset movement, as detailed in the following sections, ranged from $0.11$ to $0.13~m^3$. The larger volume of the calibration trajectory compared to the experimental trajectories is advantageous, as it minimizes errors across most tested poses. The relatively small volume for the experimental trajectories is due to the limited range along the z-axis, a result of the user's seated position.

\subsection{Evaluation of Real-World Headset Movement}
Using the same mounting setup, i.e. with the TCP-to-Unity transformation and their respective errors remaining identical, we executed the robot's trajectory, resulting in the HMD movements recorded in Section \ref{Head_Movement}. The movement was recorded during one full gameplay of \textit{Space Pirate Trainer DX}, which lasted about two minutes. Finally, we evaluated the trajectories for Quest 2 and Quest Pro and compared the poses of TCP during two executions. 

\begin{figure}[h!]
    \centering
    \includegraphics[width=0.98\columnwidth]{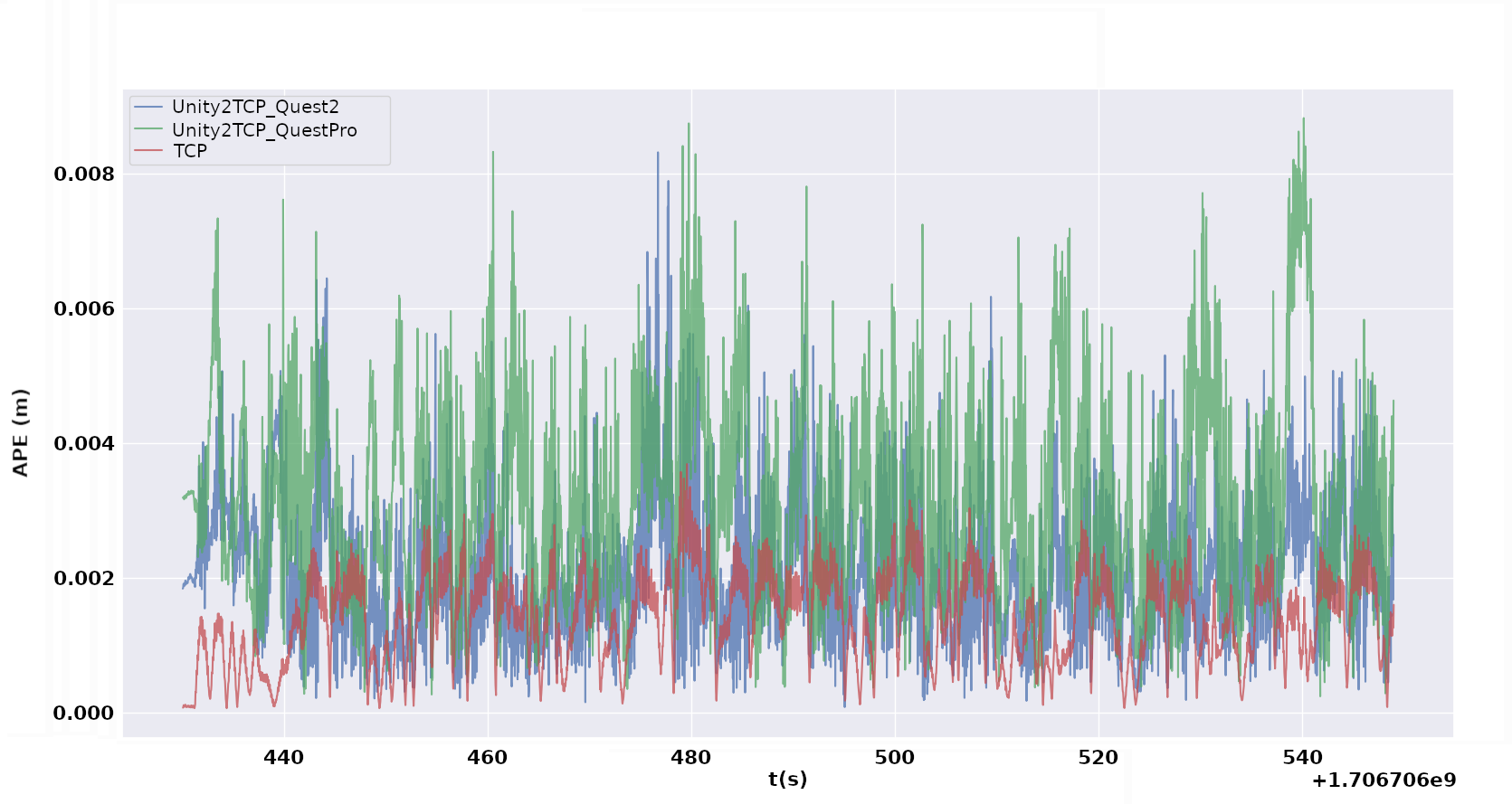}
    \includegraphics[width=0.98\columnwidth]{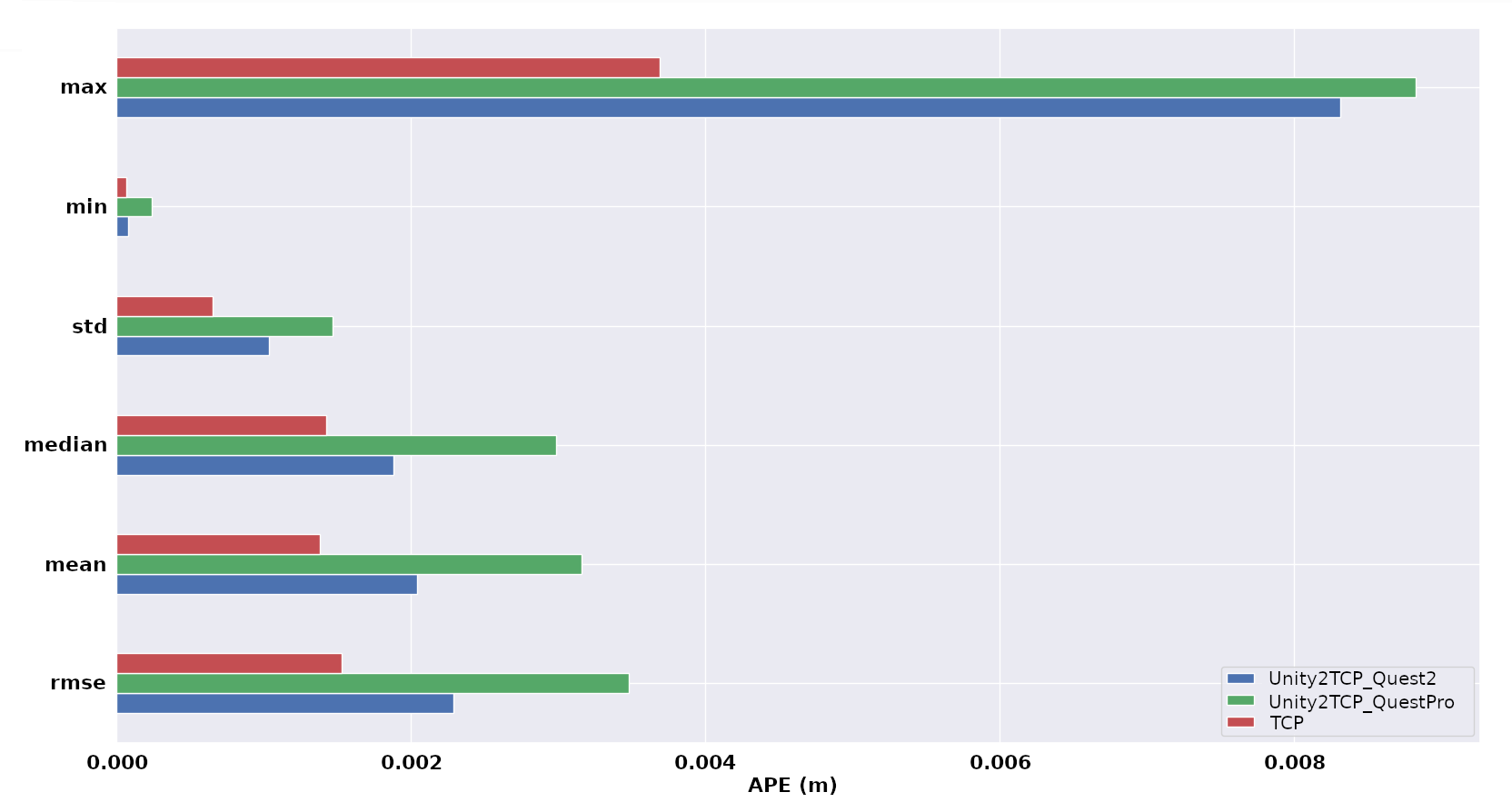}
    \caption{APE plot (up) and diagram (down) for Quest 2, Quest Pro, and TCP trajectories between two identical executions of the robot's predefined motion plan resulting in Real-World headset movements, with Unix timestamps on the plot abscissa.}
    \label{fig:APE_traj1}
%    \vskip -2mm
\end{figure}

Firstly, we noticed that the accuracy of trajectory reproduction for TCP poses in case of much more complicated and faster trajectories deteriorated. In such a scenario, the RMSE equaled $1.53[mm]$, which is almost a threefold increase as compared to the calibration sequence. The maximum error also increased significantly to $3.69[mm]$. As for Quest 2, the mean absolute pose RMSE amounted to $2.29[mm]$, and the mean error equaled $2.05[mm]$ with the maximum error reaching $8.32[mm]$. In the case of Quest Pro, RMSE was equal to $3.49[mm]$, and the mean error leveled at $3.16[mm]$ with the maximum error reaching $8.83[mm]$ (see Fig.~\ref{fig:APE_traj1}).

\begin{figure}[h]
\includegraphics[width=1\columnwidth]{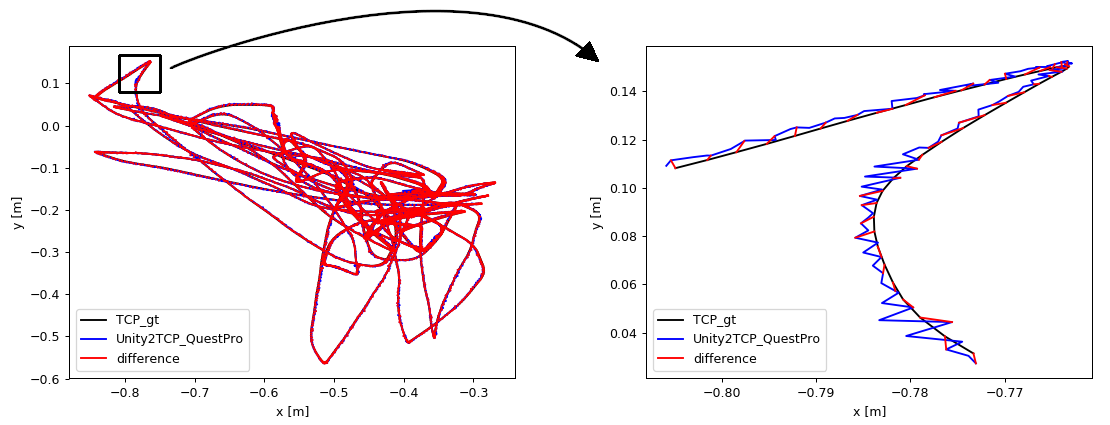}
% \caption{The Absolute Trajectory Error (ATE) plot for the real-world Quest Pro headset trajectory reproduced on a robot.}
\caption{The APE plot for a real-world Quest Pro headset trajectory reproduced on a robot. Errors are very small and only visible in the close-up due to the scale.}
\end{figure}

The final positioning accuracy results for these two headsets were comparable, with a slight indication in favor of the Quest 2, which outperformed Quest Pro in every studied metric except for minimal error. 

However, the differences between the headsets were within the limits of calibration accuracy. Therefore, it is difficult to assess unambiguously whether Quest 2 is more accurate. Instead, we can conclude that both headsets maintain high accuracy throughout the real-world trajectory without a single error of $9.0[mm]$.

We repeated the experiment to verify the repeatability of the systems' performance, yielding the results shown in Tab.~\ref{table:results_repeated_traj1}. As we can observe, the behaviors and relationships between the results are consistent, exhibiting only slightly elevated error values.

\begin{table}[h!]
\begin{tabularx}{\columnwidth}{Xcccc}
\hline
\textbf{Trajectory} & \textbf{RMSE} & \textbf{Mean} & \textbf{Std} & \textbf{Max} \\
\hline
Meta Quest 2     & 3.19 & 2.85 & 1.46 & 9.51 \\
Meta Quest Pro & 5.29 & 4.74 & 2.34 & 13.48 \\
TCP       & 2.40 & 2.16 & 1.05 & 5.11 \\
\hline
\end{tabularx}
\caption{Results for repeated experiment with the real-head trajectory on the robot provided in $[mm]$.}
\label{table:results_repeated_traj1}
\end{table}

In the analyzed trajectory, we conducted separate tests in which we obstructed two and three out of five RGB cameras on the Quest Pro, each responsible for capturing the surrounding environment.
Obscured were successively RGB cameras with id $[0,3]$ and $[0,2,3]$.
The results, presented in Tab.~\ref{table:results_covered}, indicate a slight decrease in tracking accuracy when cameras were obscured. However, covering 2 and 3 cameras yielded comparable results.
During this experiment, the hand-eye calibration error was $2.8[mm]$ for translation and $0.179\degree$ for rotation.

\begin{table}[h!]
\begin{tabularx}{\columnwidth}{Xcccc}
\hline
\textbf{Trajectory} & \textbf{RMSE} & \textbf{Mean} & \textbf{Std} & \textbf{Max} \\
\hline
0 cameras obstructed   & 3.05 & 2.69 & 1.44 & 8.73 \\
2 cameras obstructed   & 6.08 & 5.49 & 2.62 & 13.60 \\
3 cameras obstructed   & 5.56 & 5.06 & 2.39 & 13.59 \\
\hline
\end{tabularx}
\caption{Evaluation of the same trajectory with different numbers of obstructed cameras.}
\label{table:results_covered}
\end{table}

\subsection{Evaluation of Long-Period Movement Trajectory}
We also examined the accuracy of the headset over a more extended period in real-world conditions. With an exemplary head trajectory recorded during gameplay, we decided to loop it for two hours, simulating an extended gaming session with the Quest Pro.

\begin{figure}[h!]
    \centering
    \includegraphics[width=0.98\columnwidth]{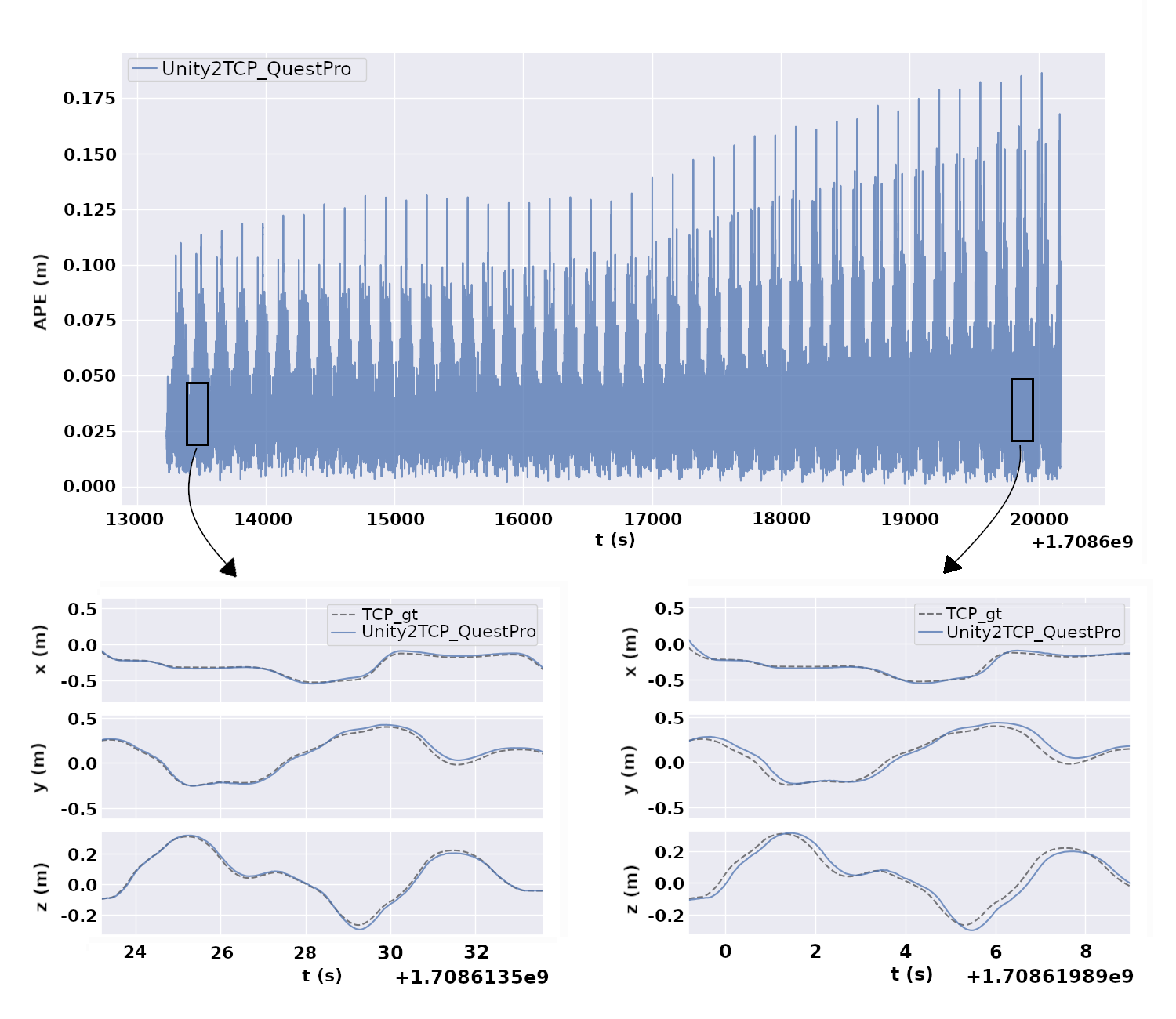}
    \caption{The APE plot during a two-hour-long looped trajectory with corresponding pieces of axial trajectory from the beginning and the end of the experiment with Unix timestamps displayed on the abscissa.}
    \label{fig:2h_traj_t_offset}
%\vskip -4mm
\end{figure}

This test was conducted on a different date than the previous ones, requiring us to first perform another calibration procedure. Its results this time were not so accurate, with a $42.74[mm]$ translation error and $3.96\degree$ rotation error. This deterioration was likely caused by an accidental movement of one of the system components. However, for this experiment, the primary objective was to examine the error variation over time. Therefore, we deemed the calibration sufficient for this purpose and proceeded, bearing in mind that values on the ordinate were considerably high.

The visible periodic spike in the pose error is caused by the robot's brake, which occurred during a short moment of reloading the looped trajectory to the controller at its end.

Throughout the experiment, the time offset between the robot and Unity data increased, directly deteriorating the results seen on the APE plot shown in Fig.~\ref{fig:2h_traj_t_offset}. This also confirms the necessity of our peak detection method, which helps level up this phenomenon. This demonstrates that evaluating longer trajectories recorded in Unity with external ground truth is particularly challenging due to the progressively increasing divergence of their timestamps.

\section{Conclusion}
The presented system for evaluating the accuracy of VR headset positioning, which integrates motion capture and robotic trajectory reproduction, demonstrated high precision in both the collection and reproduction of headset movements. Utilizing advanced spatial calibration and temporal synchronization techniques, we achieved reliable reproduction of real-world VR trajectories.

Our findings indicate that the Meta Quest 2 and Quest Pro headsets exhibit similar levels of accuracy, with the Quest 2 showing a slight advantage in some metrics. The fact that the more affordable Quest 2 matches the accuracy of the more expensive Quest Pro can provide important insight for developers, including those in industrial VR applications.
The extended evaluation revealed challenges in time synchronization between the robot and Unity data, underscoring the need for advanced synchronization methods in long-term VR accuracy assessments.

Our approach provides a comprehensive and precise method for assessing VR device's positioning accuracy, which is crucial for enhancing VR experiences and developing more accurate immersive applications. It appears to be the first of its kind to test the accuracy of different headsets on real pre-recorded trajectories and can be easily reproduced from commodity components.
Further research can use a motion capture system to independently assess the accuracy of the trajectories replicated by the industrial robot.
This shall minimize the influence of any intrinsic calibration issues in the robot arm on the results, but at the expense of a more complicated experimental setup. 
Moreover, conducting experiments for headsets that are provided by vendors other than Meta and use localization based on different hardware and visual odometry algorithms is an obvious extension.

% \section*{Supplemental Materials}
% \label{sec:supplemental_materials}

% Refer to the instructions for this section (\cref{sec:supplement_inst}).
% Below is an example you can follow that includes the actual supplemental material for this template:

% All supplemental materials are available on OSF at \url{https://doi.org/10.17605/OSF.IO/2NBSG}, released under a CC BY 4.0 license.
% In particular, they include (1) Excel files containing the data for and analyses for creating \cref{tab:vis_papers} and \cref{fig:vis_papers}, (2) figure images in multiple formats, and (3) a full version of this paper with all appendices.
% Our other code is intellectual property of a corporation---Starbucks Research---and there is no feasible way to share it publicly.

\acknowledgments{
This work was done at the Pozna\'n University of Technology Centre for Artificial Intelligence and Cybersecurity. 
The study has been supported by funding provided through an unrestricted gift by Meta. 
}

\bibliographystyle{abbrv-doi}

\balance
\bibliography{main}
\end{document}